\documentclass[fleqn,10pt]{wlscirep}
\usepackage[utf8]{inputenc}
\usepackage[T1]{fontenc}
\usepackage{lineno}
\usepackage{comment}

\title{Expert-Annotated Embryo Image Dataset with Natural Language Descriptions for Evidence-Based Patient Communication in IVF}

\author[1,$\dag$,*]{Nicklas Neu}
\author[2]{Thomas Ebner}
\author[3]{Jasmin Primus}
\author[1]{Bernhard Schenkenfelder}
\author[1]{Raphael Zefferer}
\author[3]{Mathias Brunbauer}
\author[1,$\dag$,*]{Florian Kromp}
\affil[1]{Software Competence Center Hagenberg, Hagenberg, 4232, Austria}
\affil[2]{Kepler Universitätsklinikum, Kinderwunsch Zentrum, Linz, 4020, Austria}
\affil[3]{Wunschkind Klinik Dr. Brunbauer, Wien, 1010, Austria}

\affil[*]{corresponding authors}
\affil[$\dag$]{these authors contributed equally to this work}

\begin{abstract}

Embryo selection is one of multiple crucial steps in in-vitro fertilization (IVF), commonly based on morphological assessment by clinical embryologists. Although artificial intelligence methods have demonstrated their potential to support embryo selection by automated embryo ranking or grading methods, the overall impact of AI-based solutions is still limited. This is mainly due to the required adaptation of automated solutions to custom clinical data, reliance on time lapse incubators and a lack of interpretability to understand AI reasoning. Moreover, the modern, informed patient is questioning expert decisions, particularly if the treatment is not successful. Thus, evidence-based decision justification in tasks like embryo selection would support transparent decision making and respectful patient communication, for the benefit of both, patients and experts. To support this aim, we hereby present an expert-annotated dataset consisting of embryo images previously published and corresponding morphological description using natural language. The description contains relevant information on embryonic cell cycle, developmental stage and morphological features. This dataset enables the finetuning of modern foundational vision-language models to learn to describe embryo morphology using natural language. Predicted embryo descriptions can then be leveraged to automatically extract scientific evidence from literature through advanced solutions such as retrieval-augmented generation systems or AI agent-based access, facilitating well-informed, evidence-based decision-making. This, in turn, enables transparent and justified communication with patients, ultimately enhancing the decision-making process. Moreover, language-grounded vision models are known to be a solid basis for further data-efficient, downstream task training. Our proposed dataset of natural-language human embryo image annotations is thus designed to support research in language-based, interpretable, and transparent automated embryo assessment.
\end{abstract}

\begin{document}

\flushbottom
\maketitle

\thispagestyle{empty}

\section*{Background \& Summary}

Medical Assisted Reproduction (MAR), and in particular in-vitro fertilization (IVF) and intracytoplasmic sperm injection (ICSI), have become a cornerstone of modern reproductive medicine, offering a pathway to parenthood for couples and individuals affected by infertility. The global impact of MAR is substantial, with the World Health Organization estimating that approximately 17.5\% of people of reproductive age experience infertility during their lifetime \cite{WHO}. This prevalence underscores the growing reliance on MAR as a critical intervention in reproductive healthcare. Despite remarkable progress in ovarian stimulation protocols, embryo culture systems, and laboratory instrumentation, success rates remain modest\cite{gleicher2019worldwide}. The probability of achieving a live birth per treatment cycle is around 45\%, depending on factors such as maternal age, oocyte quality, and the clinical standards of the fertility center \cite{Bhide2024}. Because the likelihood of success in a single cycle is limited, a significant proportion of patients must undergo multiple IVF cycles before achieving pregnancy. Each additional cycle imposes considerable financial costs, often not fully covered by healthcare systems, as well as physical strain from hormonal treatments and oocyte retrieval procedures. Alongside the physiological challenges, patients frequently experience psychological stress, anxiety, and emotional fatigue, which can negatively affect treatment adherence and overall well-being \cite{Chachamovich2010}. 

A crucial step determining the success of IVF is the selection of one out of multiple embryos for transfer. Commonly, embryologists rely on morphological assessment of embryo compartments such as blastocoel expansion, inner cell mass (ICM) quality and trophectoderm (TE) integrity. This assessment, guided by grading systems such as the Gardner scoring system\cite{Gardner2000BlastocystScore}, are inherently subjective, based on long-term experience and are prone to inter- and intra-observer variability. 
To enable automated embryo assessment, supervised computer vision methods such as deep learning models can be trained to learn subtle morphological cues that are often imperceptible to the human eye, leading to models' output potentially providing more consistent and objective predictions of embryo status. Recent advances in artificial intelligence (AI) have started to transform clinical embryology by enabling automated, data-driven embryo assessment complementing traditional morphological scoring\cite{Enatsu2022,Boucret2025,Kalatehjari2025,Khosravi2019,Thirumalaraju2021,Wang2021,Raef2019,Goyal2020,Cordeiro2025}. Although these methods have shown their potential, the impact on relevant performance parameters within the IVF process still needs to be demonstrated. As major drawback, the insufficient explainability of AI models hinders effective patient communication and undermines patient trust in AI-assisted decision-making or decision-support within the IVF process. This in particular holds true, as modern patients have access to large-scale literature based on internet research or due to talking to AI bots such as ChatGPT\cite{openai_chatgpt_2026} or Gemini \cite{google_gemini_2026}, and thus, patients can critically question experts choices. This is not only true for IVF treatment, but for all fields of healthcare. Since IVF treatment can be costly, stressful and is associated with high expectations in general, non-transparent or poorly-communicated decisions lead to disapproval by patients \cite{assaysh2023women, borghi2021patient}. This in particular holds true in case of cycles that have not been successful. Supporting decisions such as selecting a favorable embryo over multiple other embryos with scientific evidence and fostering transparent communication would immensely increase patient satisfaction \cite{lee_embryo_selection_challenges_2024}, but automated solutions do not exist yet.

Lately, the integration of multi-modal approaches, combining visual data with clinical information or natural language description, has gained increasing attention within the broader medical imaging domain \cite{Liu2025,Ryu2025}. 
While such methods have shown promise in improving explainability and clinical interpretability, their application to embryo assessment and IVF/ICSI remains  unexplored, highlighting a significant opportunity for innovation in this field. Of special interest are large pre-trained vision-language models, trained on billions of paired image and text data, learning to relate image features with semantic information from the natural language descriptions. It has been demonstrated that these models can be finetuned with limited, custom data such that they learn to interpret and describe images using natural language, also called image captioning. Finetuning such models could lead to AI models capable of describing embryo morphology using natural language. These descriptions could then be used by plain retrieval augmented generation (RAG) systems or could be passed to AI agents with access to RAG systems to derive relevant information from scientific literature or action recommendations such as the Istanbul Consensus\cite{istanbulconsensus}, and could support a transparent, evidence-guided decision making and patient communication strategy. However, to enable the finetuning of foundation models, expert-annotated datasets holding image-caption pairs are required, which are not publicly available for embryo imaging data.

To overcome these limitations, we hereby introduce a comprehensive dataset annotation designed specifically for research on automated and interpretable embryo assessment. In contrast to existing resources, our dataset contains natural language descriptions (captions) of morphological features, embryonic cell cycle, developmental stage, and clinically relevant observations. This combination of visual and linguistic information facilitates the development and benchmarking of vision–language models (VLMs) for embryo image analysis, enabling explainable AI systems that can generate both morphological patterns and  descriptive assessments. When aligned with scientific evidence automatically retrieved from RAG systems, embryologists' decisions could be supported and well-communicated to patients.


\section*{Methods}
\subsection*{Participants and Ethics}
This study utilizes a publicly available embryo time lapse dataset \cite{timelapsedataset}, which was previously collected and made available for research purposes. As the dataset was anonymized and de-identified, no additional ethical approval was required for this study. 
\subsection*{Data Acquisition}
The image data used in this study is a subset of 1,100 images (frames) from 704 embryo development videos\cite{timelapsedataset}, each video recorded across seven focal planes. The data was acquired using the EmbryoScope™ time lapse incubator system, which continuously monitors embryo growth in-vitro and captures  morphokinetic events during preimplantation development.

\subsection*{Data Annotation}
\begin{figure}[ht]
\centering
\includegraphics[width=\linewidth]{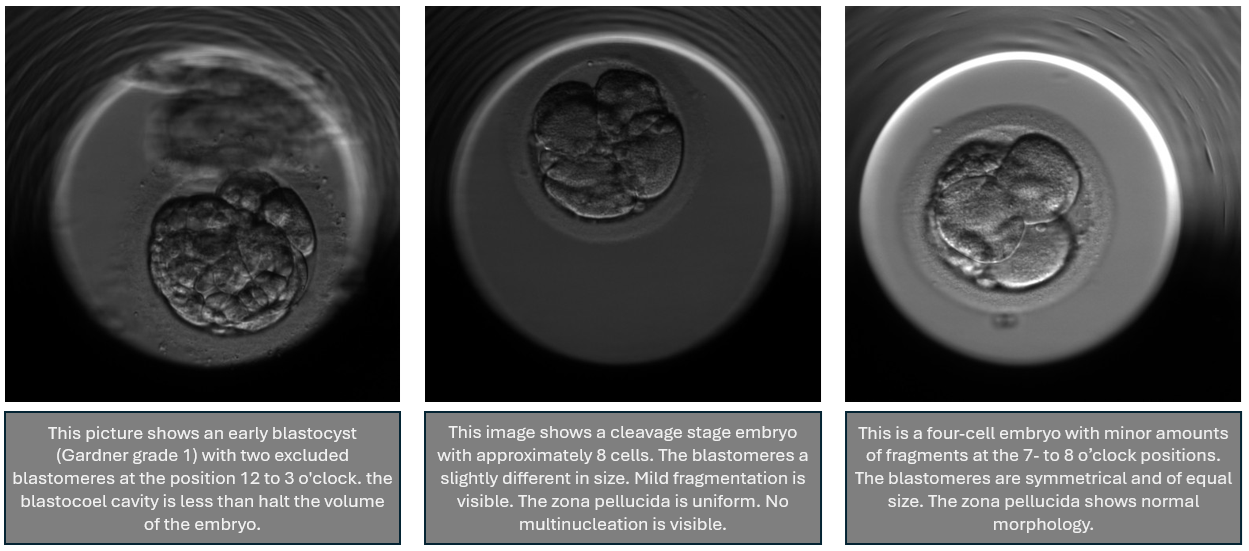}
\caption{Examples of embryo images and corresponding annotated captions.}
\label{fig:examples}
\end{figure}

Each image was annotated by trained clinical embryologists with a natural language description of embryo morphology. To this end, a web-based annotation platform was developed, built on top of the Label Studio \cite{Label_Studio} frontend. The tool enabled experts to provide a qualitative assessment (caption) of developing embryo morphology. These captions describe embryonic cell cycle, developmental stage and morphological features relevant to embryo assessment, using the language and reasoning typically used in clinical embryology practice, and following a standardized terminology aligned with established reporting conventions\cite{istanbulconsensus}. In more detail, visual attributes related to embryonic cell cycle, developmental stages (oocyte, zygote, cleavage stage, morula, blastocyst) and morphological features (e.g., polar body, perivitelline space, zona pellucida, pronuclei, cytoplasm, cell number and equality, fragmentation, compaction, expansion of blastocyst, inner cell mass, and trophectoderm quality) are described. This results in an interpretable linguistic representation of embryo morphology that closely mirrors expert evaluation. Examples of embryo images and corresponding captions are demonstrated in Figure \ref{fig:examples}.

\subsection*{Data Quality Assurance}
All annotators received detailed written instructions and example references outlining the annotation protocol and correct annotation tool usage. Annotation guidelines were derived from the Istanbul Consensus on embryo assessment published by the European Society of Human Reproduction and Embryology (ESHRE) \cite{istanbulconsensus}, which defines morphological criteria for the identification of major embryo structures.

To ensure high annotation quality and consistency, a multi-step quality assurance process was implemented. All contributors underwent comprehensive training supervised by experienced embryologists to promote consistency across annotations. Initial annotations were performed by trained  embryologists, whose work underwent review and verification by a senior clinical embryologist. In cases where uncertainties emerged during annotation, annotators consulted another senior clinical embryologist for clarification. Senior embryologists as well as undergraduate annotators were further reviewed on a random sampling basis, while all captions of the test set were reviewed by senior embryologists. This combination of structured guidance, reference examples, and expert oversight was essential to maintain the reliability and reproducibility of the morphological labels within the dataset.

\section*{Data Records}
The dataset comprises 1,100 embryo images, provided in JPEG (JPG) format. In addition, two JSONL files are distributed, containing natural language annotations (captions) forming a gold-standard training set and a silver-standard test set. The captions, authored by senior clinical embryologists, link each image to its corresponding clinical observations and morphological assessments. The gold-standard set comprises 100 images, independently annotated by both junior and senior embryologists and subsequently reviewed and consolidated by senior clinical embryologists to ensure maximal accuracy and consistency. The silver-standard set consists of the remaining 1,000 images, which were annotated by junior and senior embryologists and subjected to targeted expert review, with more than 10\% of the samples undergoing senior validation. This two-tier annotation strategy balances annotation quality and scalability, providing a highly reliable reference subset alongside a larger, carefully curated training corpus. 

\begin{figure}[ht]
\centering
\includegraphics[width=\linewidth]{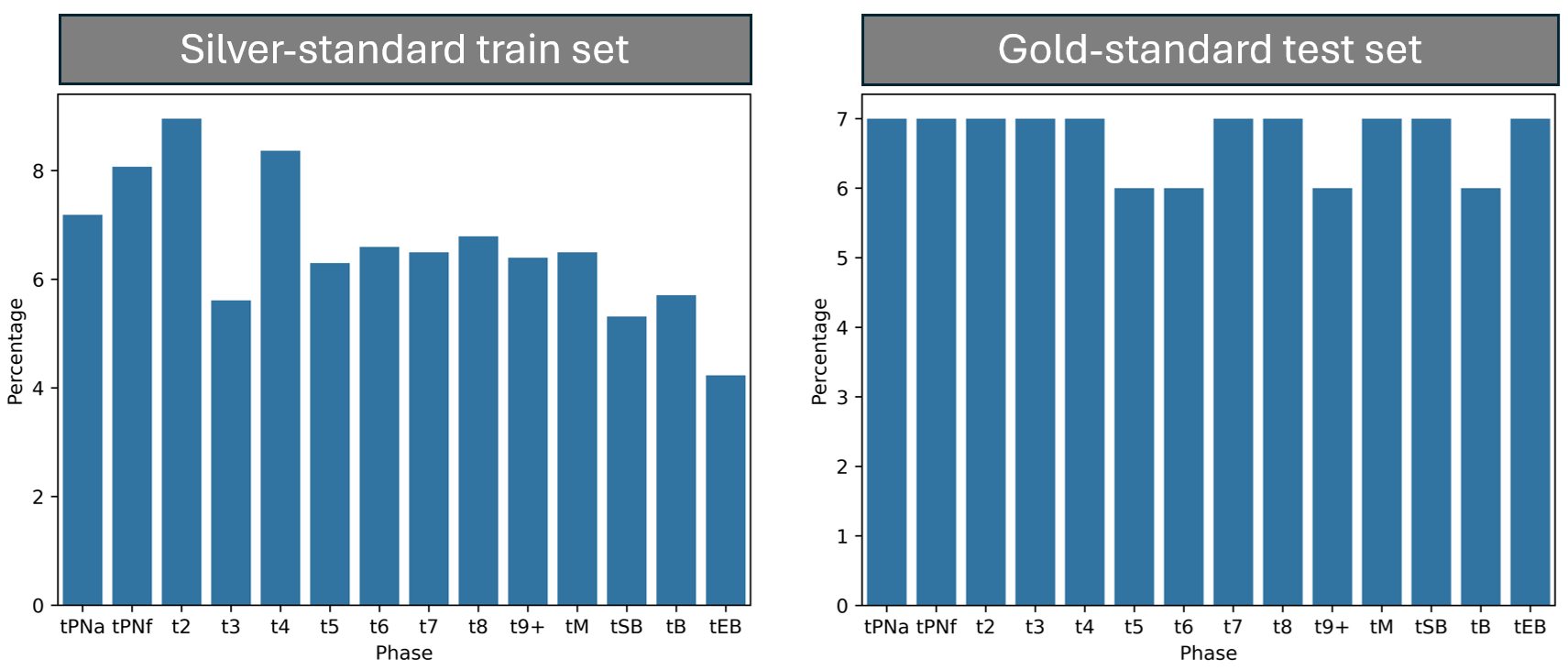}
\caption{Distribution of embryonic cell cycles in the gold- and silver-standard dataset. }
\label{fig:labels}
\end{figure}

Figure \ref{fig:labels} illustrates the relative distribution of annotated embryo images across developmental phases for the gold-standard and silver-standard caption sets, respectively. In the gold-standard set, the distribution across stages is uniform, reflecting the deliberate selection of representative samples to ensure balanced expert validation. In contrast, the silver-standard set exhibits greater variability across developmental phases, consistent with its larger size and broader coverage of routine clinical annotations.

Across both sets, later developmental stages, including morula and blastocyst-related phases remain well represented, supporting downstream analyses that rely on morphological diversity. The observed differences between the two distributions highlight the complementary roles of the gold-standard subset as a high-confidence reference and the silver-standard subset as a scalable resource for model training.

\section*{Technical Validation}

In order to guarantee the reliability and consistency of the annotated dataset, a multi-step quality assurance process was implemented as discussed in Section \textit{Data Quality Assurance}. 
To demonstrate the feasibility of training a model predicting captions from embryo images using this dataset, a recent vision-language model has been finetuned and evaluated, thereby achieving high-quality caption prediction performance \cite{kromp2026b}.

\section*{Data availability}
The dataset generated and described in this study is publicly available in a Figshare repository \cite{Kromp2026}. The repository includes embryo images in JPEG format and caption annotations provided as JSONL files (training and test splits). 
All images and annotations have been fully anonymized, and no personally identifiable information is included. The dataset is made available for research purposes in accordance with the repository’s licensing terms. More information about the research project driving this study can be retrieved \cite{birthai}. 

\bibliography{sample}

\section*{Acknowledgements} 

Supported by the Austria Wirtschaftsservice Gesellschaft mbH (aws), using funds from the National Foundation for Research, Technology and Development (Future Austria Fund) under the Grant No. P2508804.

\section*{Author contributions statement}

N. Neu performed all model trainings, experiments, and evaluations. R. Zefferer and B. Schenkenfelder created the tool to collect expert annotations. T. Ebner and J. Primus provided their expert knowledge by annotating the dataset and proofreading the manuscript. M. Brunbauer supported data annotation and proofread of the manuscript. F. Kromp coordinated the project, contributed in writing and proofreading of the manuscript. The AI-based Nature Research Assistant was used for proofreading and to make suggestions of possible improvements or how to shorten paragraphs. A self-hosted open-source large language model \cite{Llama-4-Scout-17B-16E} was used to suggest sentence rephrasing.

\section*{Competing interests} 

The authors declare no competing interests.

\section*{Figures \& Tables}

\begin{figure}[ht]
\centering
\includegraphics[width=\linewidth]{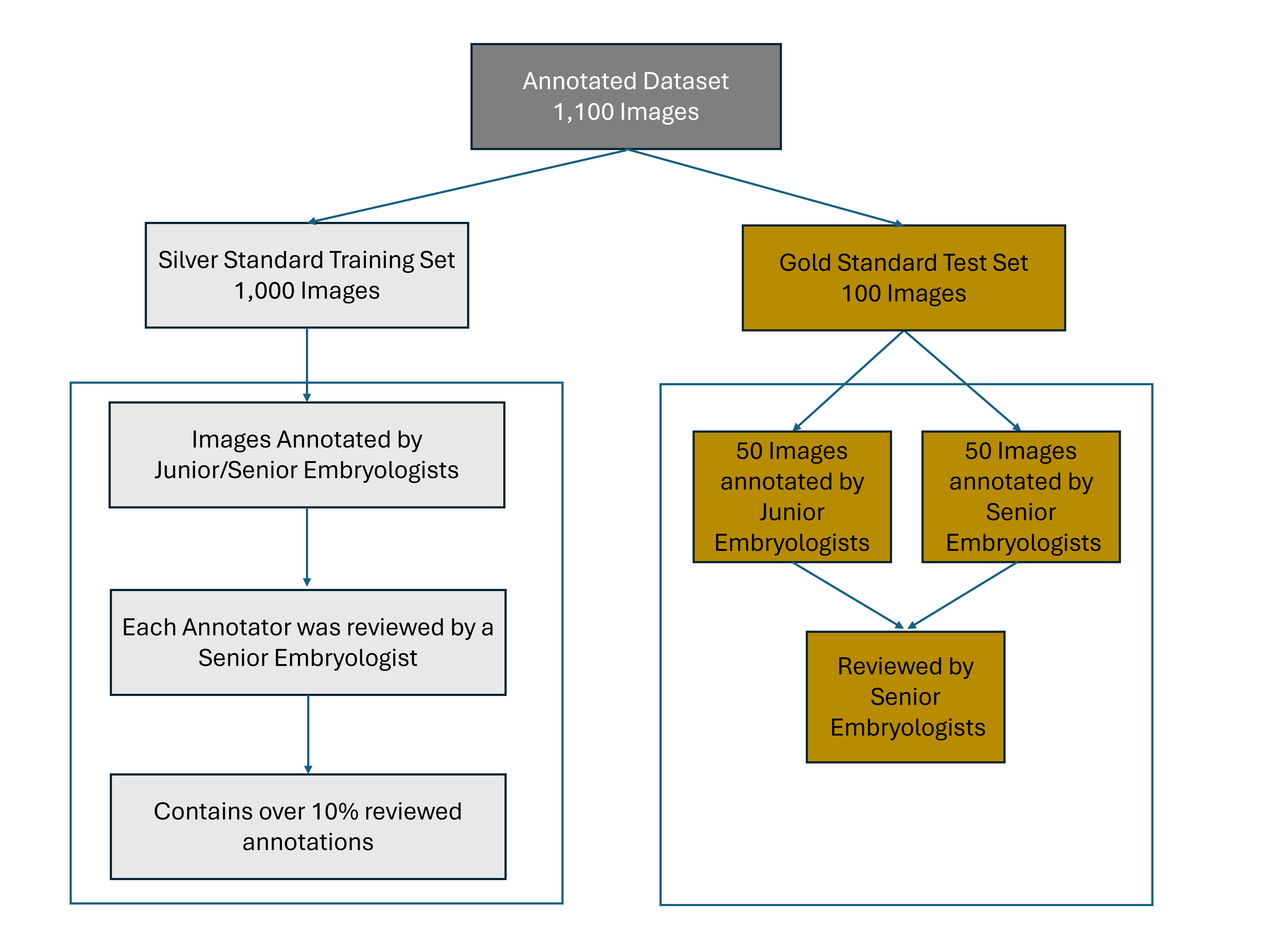}
\caption{The annotated dataset comprises 1,100 embryo images and is divided into a gold-standard and a silver-standard subset. The gold-standard set includes 100 images independently annotated by junior and senior embryologists and subsequently reviewed by senior experts to ensure maximal annotation accuracy and consistency. The silver-standard set consists of 1,000 images annotated by junior and senior embryologists, with more than 10\% of the annotations undergoing senior expert review. This two-tier annotation strategy balances annotation quality and scalability for downstream multimodal learning and evaluation tasks.}
\label{fig:structure}
\end{figure}

\end{document}